\pdfoutput=1
\documentclass[11pt]{article}

\usepackage[final]{acl}

\usepackage{times}
\usepackage{latexsym}

\usepackage[T1]{fontenc}
\usepackage[utf8]{inputenc}

\usepackage{microtype}

\usepackage{inconsolata}

\usepackage{graphicx}

\newcommand{\ours}{\textsc{ANGEL}}
\newcommand{\Ours}{Le\textbf{a}rning from \textbf{N}egative Samples in  Biomedical \textbf{G}enerative \textbf{E}ntity \textbf{L}inking}

\usepackage{booktabs}
\usepackage{caption}
\usepackage{tabularx}
\usepackage{array}
\usepackage{multirow}
\usepackage{amsmath}
\usepackage{amsfonts}

\usepackage{xcolor}
\usepackage{pifont}
\newcommand{\cmark}{\ding{51}}%
\newcommand{\xmark}{\ding{55}}%
\title{~\Ours{}}

\author{
  \textbf{Chanhwi Kim\textsuperscript{1*}},
  \textbf{Hyunjae Kim\textsuperscript{1*}},
  \textbf{Sihyeon Park\textsuperscript{1}},
  \textbf{Jiwoo Lee\textsuperscript{1}},
\\
  \textbf{Mujeen Sung\textsuperscript{2$\dagger$}},
  \textbf{Jaewoo Kang\textsuperscript{1,3$\dagger$}}
\\
\\
  \textsuperscript{1}Korea University,
  \textsuperscript{2}Kyung Hee University,
  \textsuperscript{3}AIGEN Sciences
\\
  \texttt{\{chanhwi\_kim, hyunjae-kim, sihyeon-park, hijiwoo7\}@korea.ac.kr}
\\
  \texttt{mujeensung@khu.ac.kr}, \texttt{kangj@korea.ac.kr} 
}

\begin{document}

\maketitle

\newcommand\blfootnote[1]{%
  \begingroup
  \renewcommand\thefootnote{}\footnote{#1}%
  \addtocounter{footnote}{-1}%
  \endgroup
} 

\blfootnote{\textsuperscript{$*$} Co-first authors; \textsuperscript{$\dagger$} Co-corresponding authors}

\begin{abstract}

Generative models have become widely used in biomedical entity linking (BioEL) due to their excellent performance and efficient memory usage. 
However, these models are usually trained only with positive samples, i.e., entities that match the input mention's identifier, and do not explicitly learn from hard negative samples, which are entities that look similar but have different meanings. 
To address this limitation, we introduce~\ours{} (Le\textbf{a}rning from \textbf{N}egative Samples in Biomedical \textbf{G}enerative \textbf{E}ntity \textbf{L}inking), the first framework that trains generative BioEL models using negative samples. 
Specifically, a generative model is initially trained to generate positive entity names from the knowledge base for given input entities.
Subsequently, both correct and incorrect outputs are gathered from the model's top-k predictions.
Finally, the model is updated to prioritize the correct predictions through preference optimization.
Our models outperform the previous best baseline models by up to an average top-1 accuracy of 1.4\% on five benchmarks.
When incorporating our framework into pre-training, the performance improvement increases further to 1.7\%, demonstrating its effectiveness in both the pre-training and fine-tuning stages.
The code and model weights are available at~\url{https://github.com/dmis-lab/ANGEL}.

\end{abstract}

\section{Introduction}

\begin{figure}[t!]
    \includegraphics[width=\linewidth]{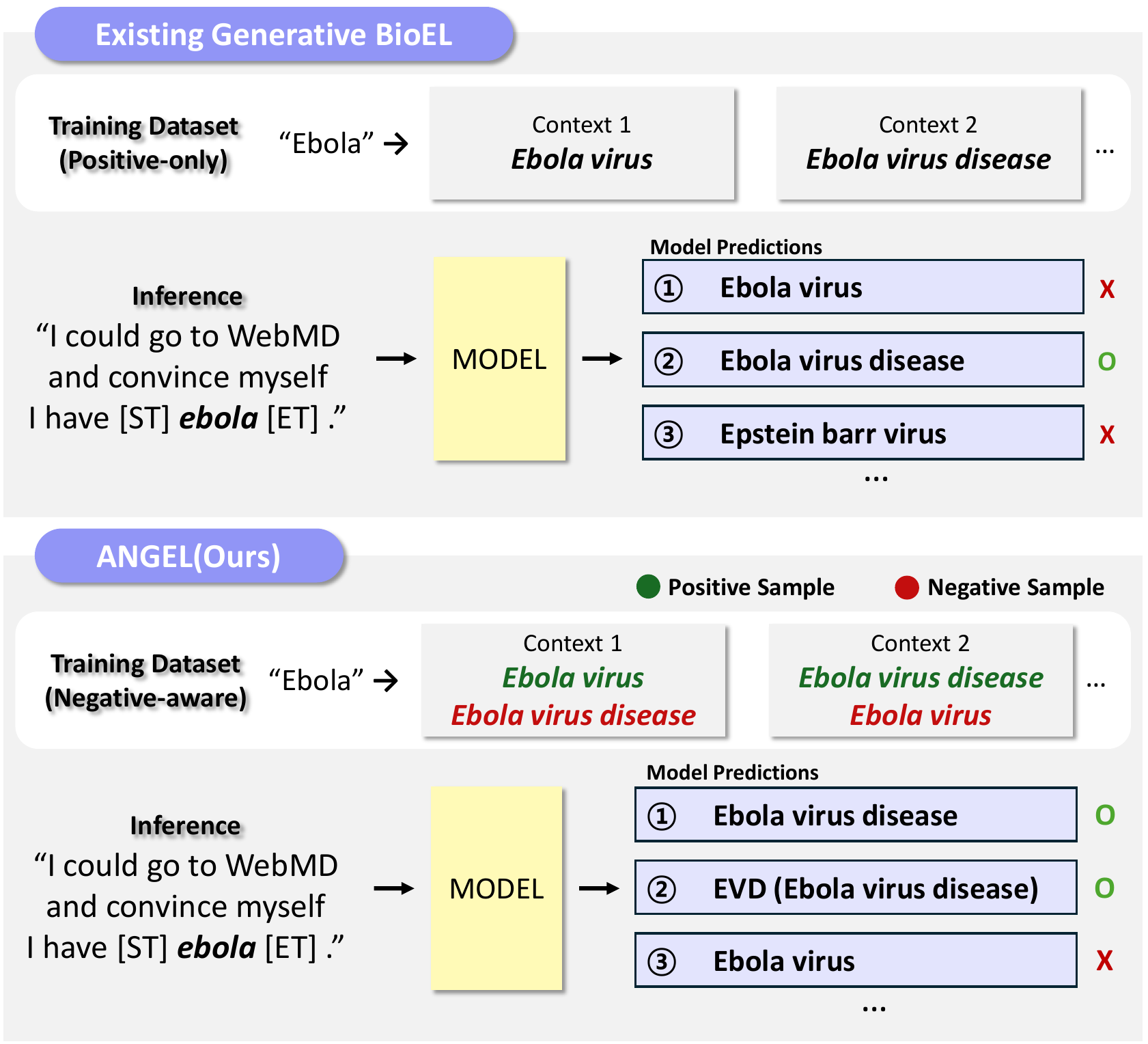}
    \caption{
    Comparison of training approaches between existing generative BioEL models and our~\ours{} method. 
    The main limitation of current generative BioEL methods is that they are trained only on positive samples. 
    This restricts their ability to distinguish between entity names that are similar in surface form but different in meaning depending on the context. 
    Our~\ours{} framework addresses this issue by training the model to prefer positive samples over negative ones.
    }
\label{figure:sub_figure}
%\vspace{-5mm}
\end{figure}

% mention  : hemiplegic migraine
% genbioel : hemiplegias
% angel    : hemiplegic ophthalmoplegic migraine

% mention  : r-aminobutyric acid
% genbioel : aminobutyric acid
% angel    : aminobutyric acid 04

% mention  : malate
% genbioel : maleate
% angel    : malate

% mention  : hemorrhagic bronchopneumonia
% genbioel : intracranial hemorrhage
% angel    : bronchopneumonia

Biomedical entity linking (BioEL) involves aligning entity mentions in text with standardized concepts from biomedical knowledge bases (KB) such as UMLS~\cite{umls} or MeSH~\cite{mesh}.\footnotemark[1]
BioEL encounters significant challenges due to the diverse and ambiguous nature of biomedical terminology, including synonyms, abbreviations, and terms that look similar but have different meanings.
For instance, `ADHD' (CUI:{\textit{C1263846}}, where CUI stands for Concept Unique ID) has synonyms such as hyperkinetic disorder and attention deficit hyperactivity disorder. 
Additionally, `ADA' can be mapped to either adenosine deaminase (CUI:{\textit{C1412179}}) or American Diabetes Association (CUI:{\textit{C1705019}}) depending on the context in which the entity appears. 

\footnotetext[1]{%
  UMLS and MeSH are short for the Unified Medical Language System and Medical Subject Headings, respectively.
}

Recent studies have focused on addressing these challenges, broadly categorized into two approaches: similarity-based and generative BioEL.
Similarity-based models~\cite{biosyn, sapbert, rescnn, bhowmik2021fast} encode input mentions and entities from KBs into the same vector space using embedding models. 
They then calculate similarity scores to identify the most similar entities for each input entity.
Although these approaches have achieved remarkable improvements, they require significant space to index and load embedding vectors for all candidate entities~\cite{genre}.
Furthermore, representing both the input and candidate entities as single vectors using a bi-encoder can limit the quality of their representations, making it difficult to handle challenging cases.

On the other hand, generative models~\cite{genre,biobart,genbioel}, built upon an encoder-decoder structure~\cite{bart,t5}, directly generate the most likely entity name from the KB for the input entity. 
The output space is dynamically controlled through a constrained decoding strategy, ensuring that only entities from the target KB are generated.
Generative models offer several advantages over similarity-based models, including greater memory efficiency and higher performance. 
They eliminate the need to index large external embedding vectors, and their auto-regressive formulation effectively cross-encodes the input document and candidate entities. 

However, existing generative models are trained solely on positive samples and do not explicitly learn from negative samples.
Despite their high performance, they encounter limitations when distinguishing between biomedical entities with similar surface forms but different meanings.
Although similarity-based models address this issue by incorporating negative samples through synonym marginalization~\cite{biosyn} or contrastive learning~\cite{sapbert}, applying these approaches to generative models is not straightforward.
Consequently, generative models may overfit to surface-level features, reducing the models' ability to generalize effectively across varied contexts, as illustrated in Figure~\ref{figure:sub_figure}.

\begin{figure*}[t]
    \includegraphics[width=\textwidth]{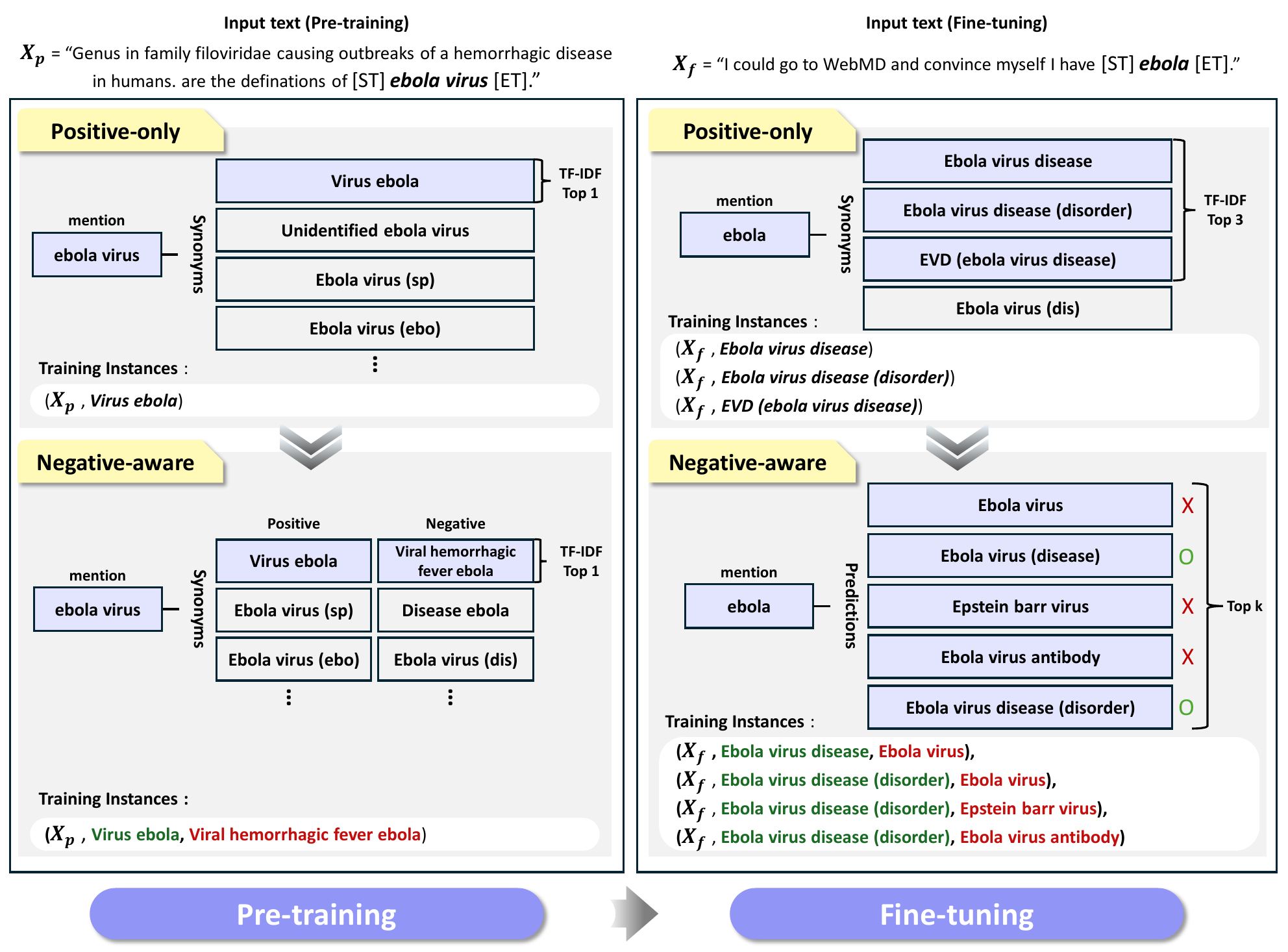}
    \caption{Overview of our method~\ours{}. 
    The core idea is to enhance both pre-training and fine-tuning by incorporating negative samples, which are obtained either through TF-IDF similarity or the model's top-k predictions.
    This approach helps the model distinguish subtle differences between correct and incorrect entities.
    }
    \label{figure:main_figure}
    %\vspace{-3mm}
\end{figure*}

To harness the benefits of generative approaches while overcoming their limitation of not using negative samples, we introduce a novel training framework,~\ours{}.
Our framework operates in two stages: positive-only training and negative-aware training (see Figure~\ref{figure:main_figure}).
In the first stage, a generative model is trained to generate biomedical terms from the KB that share the same identifier as the given input entity.
In the second stage, we gather both correct and incorrect outputs from the model's top-k predictions. 
The model is then updated to prioritize the correct predictions using a preference optimization algorithm~\cite{burges2010ranknet,DPO}.
Models trained on our \ours{} framework significantly outperform the previous best similarity-based and generative BioEL models, achieving an average accuracy improvement of 1.7\% across five datasets.
Our contributions are as follows:
\begin{itemize}
    \item We introduce~\ours{}, the first-of-its-kind training framework that utilizes negative samples in generative entity linking.
    \ours{} overcomes the limitations of existing generative approaches by effectively employing negative samples during training.
    \item \ours{} is a versatile framework, demonstrating its applicability in both the pre-training and fine-tuning phases, leading to performance improvements at each stage.
    Additionally, our method is model-agnostic, consistently improving results across various backbone language models, with gains ranging from 0.9\% to 1.7\%.
    \item Our best model, pre-trained and fine-tuned with our framework, outperforms the previous best baseline model by 1.7\% across five benchmark datasets.
\end{itemize}

\section{Related Work}

\subsection{Biomedical Entity Linking}
Biomedical entity linking (BioEL), also known as biomedical entity normalization, is a crucial task because of its application in several downstream tasks in the biomedical domain, such as literature search~\cite{lee2016best}, knowledge extraction~\cite{li2016biocreative,xiang2021ontoea,zhang2023emerging}, knowledge graph alignment~\cite{cohen2005survey,lin2022multi}, and automatic diagnosis~\cite{shi2021understanding,yuan2024efficient}.
Typically, it is assumed that the target mention is already provided, and the task is solely to link this mention to the appropriate entity name from the KB.
End-to-end BioEL~\cite{zhou2021end,ujiie2021end}, which also involves identifying mentions within a sentence, is being actively researched, but this is not our focus and will not be discussed in detail.

Traditional classification-based approaches \cite{limsopatham2016normalising, miftahutdinov2019deep} employed a softmax layer for classification, treating concepts as categorical variables and thereby losing the detailed information of concept names.
Similarity-based~\cite{biosyn, sapbert, rescnn, krissbert} models have significantly improved BioEL performance, which encodes mentions and candidate entity names in the same vector space.
They are characterized by high memory consumption due to the need to encode entities into pre-computed embeddings, posing scalability challenges with large datasets~\cite{genre}.
integrated the concept of clustering into BioEL \cite{angell2021clustering,agarwal2022entity}.

\subsection{Generative Entity Linking}
Generative models have become a powerful method for entity linking by overcoming the limitations of similarity-based models. 
The GENRE framework~\cite{genre} was the first to demonstrate this approach. 
To enhance precision and reduce memory usage, GENRE introduced a constrained decoding method~\cite{beamsearch} using a prefix tree (trie), which restricts the output space to valid entity names.
This technique also facilitates easy updates to the set of entities, making the system highly adaptable to changes in the KB.
In the biomedical field, notable examples of generative models include GenBioEL~\cite{genbioel} and BioBART~\cite{biobart}.
GenBioEL, in particular, is the first model to apply a generative model BART~\cite{bart} to BioEL, after pre-training it using UMLS.
Additionally, several hybrid approaches, known as retrieve-and-generate methods, have been proposed~\cite{promptBioEL,bioelqa}.
In these methods, a similarity-based model first retrieves the top-k candidates, which are then reranked using a generative model.
Although generative approaches have shown high performance, their training has typically been limited to positive samples, as discussed in the introduction section.
This absence of explicit negative sample learning often leads to confusion when entities share very similar surface forms but represent different concepts.
In this study, we introduce the use of negative samples during training and demonstrate that this approach can significantly enhance the performance of generative models.

\section{Method}

\subsection{Task Formulation}

Let $ \mathcal{D} = \{(\mathbf{x}_n, y_n)\}_{n=1}^N $ be a human-labeled dataset, where $\mathbf{x}_n$ represents an input text and $y_n$ is the gold identifier defined in a KB denoted by $\mathcal{E}$.
Each $\mathbf{x}_n = (\mathbf{c}^-_n, \mathbf{m}_n, \mathbf{c}^+_n)$ contains a target entity mention $\mathbf{m}_n$ along with its surrounding contextual information $\mathbf{c}^-_{n}$ and $\mathbf{c}^+_{n}$, which represents the tokens before and after the entity mention $ \mathbf{m}_{n} $, respectively.
For simplicity, we will omit the subscript $n$.
Our goal is to map each mention $\mathbf{m}$ to its corresponding gold identifier $y$ from the set of entity names $\mathcal{E}$. 
To achieve this, we define the model's prediction $y^*$ as follows:
%\mujeen{$p(;\theta) => p_{\theta}$}
\begin{equation}
    y^* = \mathcal{F}(\operatorname{argmax}_{\mathbf{e} \in \mathcal{E}} \; p_\theta(
    \mathbf{e}|\mathbf{m})), 
\end{equation}
where $\mathbf{e}$ is an entity name defined in the KB, $\mathcal{F}$ is a function that aligns entities to their identifiers, and $\theta$ represents the model parameters. 

A single gold identifier may have multiple associated entity names that refer to the same concept; we refer to these as synonyms. 
Previous generative BioEL approaches train the model to generate a textual synonym $\mathbf{s} \in \mathcal{S}_y$, where $\mathcal{S}_y \subset \mathcal{E}$ denotes the set of entity names associated with the identifier $y$, in an autoregressive manner as follows:
\begin{equation}
p_{\theta}(\mathbf{s} \mid \mathbf{x}, \mathbf{v}) = \prod_{t=1}^{T} p_{\theta}(s_t \mid s_{<t}, \mathbf{x}, \mathbf{v}),
\label{equation:positive}
\end{equation}
where $T$ is the number of tokens of the synonym $\mathbf{s}$, $s_t$ indicates the $t$-th token of the synonym, and $\mathbf{v}$ is the prompt.
In an encoder-decoder model structure~\cite{bart}, the input to the encoder is formatted as follows:
\begin{align*}    
&\texttt{[BOS]} \; \mathbf{c}^- \; \texttt{[ST]} \; \mathbf{m} \; \texttt{[ET]} \; \mathbf{c}^+ \; \texttt{[EOS]},
\end{align*}
where the special tokens \texttt{[ST]} and \texttt{[ET]} surround the target mention, and the special tokens \texttt{[BOS]} and \texttt{[EOS]} represent the `Begin Of Sentence' and `End Of Sentence,' respectively.
The prefix prompt $\mathbf{v}$ to the decoder, represented as `\texttt{$\mathbf{m}$ is}', is concatenated with \texttt{[BOS]} and input to the decoder. The prompt is designed to make the decoder's output resemble a natural language sentence, which helps to minimize discrepancies between language modeling and fine-tuning on the BioEL task.

As shown in Equation~\ref{equation:positive}, existing models are trained solely to predict synonyms for the input mention (i.e., positive samples), without leveraging negative samples. 
In contrast, we propose a novel approach using negative samples, which we will describe in detail in the following sections.

\subsection{\ours{} Framework}
Our framework comprises two main stages: positive-only training, which warms up the model using positive samples to learn morphological similarities among synonyms, and negative-aware training, which progressively refines the model by incorporating negative samples (see Figure~\ref{figure:main_figure}).

\paragraph{Positive-only training}

We initialize the model to generate synonyms, similar to previous methods (see Equation~\ref{equation:positive}).
For the input mention, we select the most similar synonyms based on their vector similarity, which is calculated as follows:
\begin{equation}
    \mathcal{\hat{S}}_y = \operatorname{argsort}_{\mathbf{s} \in \mathcal{S}_y}(\operatorname{TFIDF}(\mathbf{m}, \mathbf{s})),
\label{equation:tfidf}
\end{equation}
where $\operatorname{TFIDF}(\cdot)$ returns the TF-IDF similarity between tri-grams of the mention and its synonyms.
We use the top-k subset $\mathcal{\hat{S}}_y[:\text{k}] = \{\mathbf{\hat{s}}^\text{(1)},\dots,\mathbf{\hat{s}}^\text{(k)}\}$ as training instances for each mention.

\paragraph{Negative-aware training}

After obtaining the top-k predictions from the model for each mention in the training set, we construct a dataset of triplets $(\mathbf{x}, \mathbf{e}_w, \mathbf{e}_l)$, where $\mathbf{x}$ denotes the mention along with its context (if available),
$\mathbf{e}_w$ is the correct (preferred) entity, and $\mathbf{e}_l$ is an incorrect (dispreferred) entity.
From all possible $(\mathbf{e}_w, \mathbf{e}_l)$ pairs, we retain only those for which the model ranks the incorrect entity $\mathbf{e}_l$ above the correct one $\mathbf{e}_w$, thereby reflecting an incorrect model preference. If the top-ranked prediction is already correct, we pair this correct entity $\mathbf{e}_w$ with the highest-ranked incorrect entity $\mathbf{e}_l$ to preserve the model’s original preference structure.

We denote the resulting training set by $\mathcal{D}'$ and fine-tune the model using a pairwise preference loss, formulated as follows:
\begin{equation}
\begin{aligned}
\mathcal{L}(\theta) =& 
- \mathbb{E}_{(\mathbf{x}, \mathbf{e}_w, \mathbf{e}_l) \sim \mathcal{D}'} [
   \\ & \log \sigma (
        \beta (r_\theta(\mathbf{e}_w \mid \mathbf{x}) - r_\theta(\mathbf{e}_l \mid \mathbf{x}) ))],
\end{aligned}
\label{equation:general_loss}
\end{equation}
where $r_\theta(\mathbf{e} \mid \mathbf{x})$ is a differentiable scoring function (e.g., $\log p_\theta(\mathbf{e} \mid \mathbf{x})$), $\sigma$ is the sigmoid function, and $\beta$ is a temperature (scaling) hyperparameter.
A recent instantiation of this general preference learning framework is Direct Preference Optimization (DPO)~\cite{DPO}, where the scoring function is defined as a log-ratio with respect to a reference model $p_{\mathrm{ref}}$ as follows:
\begin{equation}
r_\theta(\mathbf{e} \mid \mathbf{x}) 
= \log \frac{p_\theta(\mathbf{e} \mid \mathbf{x})}{p_{\mathrm{ref}}(\mathbf{e} \mid \mathbf{x})},
\end{equation}
where $p_\theta$ is the generative model being trained, $p_\text{ref}$ is a reference generative model trained in a prior stage using positive-only data.
We adopt DPO as a practical instantiation within our framework because it offers a principled and empirically effective way to incorporate prior model behavior through a reference distribution.

\paragraph{Applying~\ours{} in pre-training}

Our framework supports not only fine-tuning with labeled datasets but also pre-training with the KB. 
Specifically, we automatically generate surrounding contextual information for each entity in the KB, using clause templates or definitions, as outlined in GenBioEL~\cite{genbioel}.
When the entity definition $\mathbf{d}_y$ corresponding to the identifier $y$ is available, a synonym and its definition are integrated into a pre-defined clause template as follows:\footnote{Refer to \citet{genbioel} for the full set of templates.} 
\begin{align*}
\texttt{[BOS]} \; \texttt{[ST]} \; \mathbf{s} \; \texttt{[ET]} \; \texttt{is defined as} \; \mathbf{d}_y \; \texttt{[EOS]}.
\end{align*}
When no definitions are available in the KB, we replace $\mathbf{d}_y$ with alternative synonyms as follows:
\begin{align*}
\texttt{[BOS]} \; \texttt{[ST]} \; \mathbf{s}_1 \; \texttt{[ET]} \; \texttt{has synonyms} \\ 
\texttt{such as} \; \mathbf{s}_2 \; \texttt{[EOS]},
\end{align*}
where $\mathbf{s}_1$ and $\mathbf{s}_2$ are different synonyms.
The input for the decoder is ``$\texttt{[BOS]} \; \mathbf{s }$ (or $\mathbf{s}_1$)\; $\texttt{is}$'' and the expected output is another synonym (e.g., $\mathbf{s}_2$)  selected from the remaining synonyms.

The pre-training process, like fine-tuning, is divided into two stages: positive-only training and negative-aware training. 
However, due to the typically large scale of the KB, efficiency considerations are particularly important. 
In positive-only training, rather than utilizing all possible synonym combinations within the KB, we identify, for each entity, the most similar synonym based on TF-IDF similarity and designate it as the target synonym. 
For negative-aware training, instead of selecting negatives from the model’s predictions, negative samples are selected from entities exhibiting the highest TF-IDF similarity to the input mentions but possessing distinct identifiers.

\begin{table*}[t]
    \centering
    \footnotesize
    %\resizebox{\textwidth}{!}{
        \begin{tabular}{l l l l l l|l}
        \toprule
        \textbf{Model} & \textbf{NCBI} & \textbf{BC5CDR} & \textbf{COMETA} & \textbf{AAP} & \textbf{MM-ST21pv} & \textbf{Average} \\
        \midrule
        \multicolumn{6}{l}{\textit{Similarity-based BioEL \& Re-ranking}} \\
        \midrule
        BioSYN~\cite{biosyn} & 91.1 & 93.3$^\dagger$ & 71.3 & 86.5$^\dagger$ & OOM & - \\
        SapBERT~\cite{sapbert} & 92.3 & 88.6$^\dagger$ & 75.1 & 89.0 & 50.3$^\dagger$ & 79.1 \\
        ResCNN~\cite{rescnn} & 92.4 & 94.0$^\dagger$ & 80.1 & 77.4$^\dagger$ & 55.0 & 79.3 \\
        KRISSBERT~\cite{krissbert} & 91.3 & 72.0$^\dagger$ & 80.1$^\dagger$ & 83.1$^\dagger$ & 72.2 & 79.7 \\
        Prompt-BioEL~\cite{promptBioEL} & 91.9$^\dagger$ & 94.3$^\dagger$ & \underline{82.7}$^\dagger$ & 89.7$^\dagger$ & 72.6$^\dagger$ & 86.2 \\
        \midrule
        \multicolumn{6}{l}{\textit{Generative BioEL (reported)}} \\
        \midrule
        BART~\cite{bart} & 90.2 & 92.5 & 80.7 & 88.8 & 71.5 & 84.7 \\
        BioBART~\cite{biobart} & 89.9 & 93.3 & 81.8 & 89.4 & 71.8 & 85.2 \\
        GenBioEL~\cite{genbioel} & 91.9 & 93.3 & 81.4 & 89.3 & - & - \\
        \midrule
        \multicolumn{6}{l}{\textit{Generative BioEL (reproduced)}} \\ 
        \cmidrule(lr){1-7}
        BART$^\dagger$~\cite{bart} & 90.3 & 93.0 & 80.4 & 88.7 & 70.1 & 84.5 \\
        \quad + \ours{}$_\text{FT}$ (\textbf{Ours}) & 91.4 \scriptsize{(+1.1)} & 93.6 \scriptsize{(+0.6)} & 81.3 \scriptsize{(+0.9)} & 89.5 \scriptsize{(+0.8)} & 71.2 \scriptsize{(+1.1)} & 85.4 \scriptsize{(+0.9)} \\
        \cmidrule(lr){1-7}
        BioBART$^\dagger$~\cite{biobart} & 89.4 & 93.5 & 81.3 & 89.3 & 71.3 & 85.0 \\
        \quad + \ours{}$_\text{FT}$ (\textbf{Ours}) & 91.9 \scriptsize{(+2.5)}& \textbf{94.7} \scriptsize{(+1.2)}& 82.2 \scriptsize{(+0.9)}& \underline{89.9} \scriptsize{(+0.6)}& \textbf{73.4} \scriptsize{(+2.1)} & \underline{86.4} \scriptsize{(+1.4)} \\
        \cmidrule(lr){1-7}
        GenBioEL$^\dagger$~\cite{genbioel} & 91.0 & 93.1 & 80.9 & 89.3 & 70.7 & 85.0 \\
        \quad + \ours{}$_\text{FT}$ (\textbf{Ours}) & \underline{92.5} \scriptsize{(+1.5)} & 94.4 \scriptsize{(+1.3)} & 82.4 \scriptsize{(+1.5)} & \underline{89.9} \scriptsize{(+0.6)}& 71.9 \scriptsize{(+1.2)}& 86.2 \scriptsize{(+1.2)}\\
        \quad + \ours{}$_\text{PT + FT}$ (\textbf{Ours}) & \textbf{92.8} \scriptsize{(+1.8)} & \underline{94.5} \scriptsize{(+1.4)} & \textbf{82.8} \scriptsize{(+1.9)} & \textbf{90.2} \scriptsize{(+0.9)} & \underline{73.3} \scriptsize{(+2.6)} & \textbf{86.7} \scriptsize{(+1.7)}
        \\
        \bottomrule 
        \end{tabular}	
        %}
    \caption{
    The top-1 accuracy of the models across the five BioEL datasets.
    Our \ours{} framework is applied to generative BioEL models during fine-tuning (\ours{}$_\text{FT}$) and both pre-training and fine-tuning (\ours{}$_\text{PT+FT}$). 
    `$\dagger$': the results have been reproduced. 
    `OOM': an out-of-memory error occured when using a single 80G A00 GPU.
    }
    \label{tab:main_table}
    %\vspace{-2mm}
\end{table*}

\section{Experiments}

\subsection{Datasets}
We utilized five popular BioEL benchmark datasets: NCBI-disease~\cite{ncbi}, BC5CDR~\cite{bc5cdr}, COMETA~\cite{cometa}, AskAPatient~\cite{askapatient}, and Medmentions~\cite{medmentions}, with the ST21pv subset used for Medmentions.
Due to the lack of a test set in the AskAPatient dataset, we adhered to the 10-fold evaluation protocol outlined by \citet{askapatient}. 
Also, AskAPatient dataset does not include context for the mentions.
In the following tables, NCBI-disease, AskAPatient, and Medmentions are denoted as NCBI, AAP, and MM-ST21pv, respectively. 
Refer to Appendix~\ref{appendix:datasets} for detailed descriptions and statistics.

 similarity-based models~\cite{biosyn,sapbert,rescnn,krissbert} as our baselines. 
Notably, Prompt-BioEL~\cite{promptBioEL} employs a re-ranking-based approach. In the first stage, a similarity-based model, such as SapBERT, retrieves the top-k candidate entities from the knowledge base. In the second stage, these candidates are reranked using a cross-encoder.
Although Prompt-BioEL may not be directly comparable, as it incorporates additional modules on top of existing models, we report its performance alongside for reference.
Additionally, we include the previously best-performing generative models for comparison. (1) BART-large~\cite{bart} is an encoder-decoder language model pre-trained on a general-domain corpus.
(2) BioBART-large~\cite{biobart} is the BART-large model continuously pre-trained on a biomedical-domain corpus.
(3) GenBioEL~\cite{genbioel} is initialized with the weights of the BART-large model and then pre-trained specifically for BioEL using UMLS.
We excluded several models~\cite{agarwal2022entity, bioelqa} due to the lack of publicly available code or the difficulty in reproducing their reported performance.

\subsection{Implementation Details}

Our framework was applied to each of these models during fine-tuning, referred to as \ours{}$_\text{FT}$, and during both pre-training and fine-tuning, referred to as \ours{}$_\text{PT+FT}$.
For pre-training, we utilized the 2020AA version of the UMLS database,\footnote{\url{https://www.nlm.nih.gov/research/umls/licensedcontent/umlsarchives04.html}} which comprises 3.09M entities, of which 199K concepts contain definitions.
During pre-training, we saved checkpoints every 500 steps over 5 epochs and selected the best one based on the validation sets. 
We used top-3 synonyms as positive samples in positive-only training.
The other hyperparameter configurations are detailed in Appendix~\ref{appendix:hyperparameters}.
In pre-processing, following~\citet{genbioel}, we expanded abbreviations using AB3P~\cite{ab3p}, lowercase texts, mark mention boundaries with special tokens \texttt{[ST]} and \texttt{[ET]}, and discard mentions that overlap or are missing from the target KB.g pre-training, our models were trained using eight 80G A100 GPUs for 12 hours. 
During fine-tuning, a single A100 GPU was used.

\subsection{Results}
Consistent with previous studies~\cite{biosyn,sapbert}, we used accuracy at top-1 (Acc@1) as our evaluation metric, which quantifies the percentage of mentions where the model correctly ranks the gold standard identifier as the top choice. 
To assess statistical significance, we employed bootstrapping with the same sample size as the original datasets, repeating the process 100 times, followed by a paired t-test. 
Table \ref{tab:main_table} shows that our framework consistently outperformed the performance of generative models ($p \leq 8.2e^{-22}$ for all comparisons).
Specifically, our fine-tuning method (i.e., \ours{}$_\text{FT}$) improved the Acc@1 scores of BART, BioBART, and GenBioEL by 0.9\%, 1.4\%, and 1.2\%, respectively. 
When pre-training is also applied (i.e., \ours{}$_\text{PT+FT}$) to GenBioEL, the improvement increases to 1.7\%, further highlighting the effectiveness of both pre-training and fine-tuning in \ours{}.

Similarity-based models often exhibit limited robustness, with performance varying significantly across datasets. In contrast, generative models tend to deliver more consistent results, highlighting a key strength. Among the baseline methods, the re-ranking-based model Prompt-BioEL achieves strong performance, substantially outperforming its underlying retriever, SapBERT, though at the cost of increased inference time. Notably, our ANGEL$_\text{PT+FT}$ model surpasses Prompt-BioEL across all datasets without relying on any re-ranking component, achieving an average improvement of 0.5\%.
Given this strong baseline, incorporating a re-ranking component into our model in future work may further enhance performance.

\begin{table*}[t]
    \centering
    \footnotesize
    \begin{tabular}{lccccc|c}
        \toprule
        \textbf{Model} & \textbf{NCBI} & \textbf{BC5CDR} & \textbf{COMETA} & \textbf{AAP} & \textbf{MM-ST21pv} & \textbf{Average} \\
        \midrule
        \multicolumn{6}{l}{\textit{Models \textbf{with} Negative-aware Training}} \\
        \midrule
        \ours{} (\textbf{Ours}) & 92.8 & \textbf{94.5} & \textbf{82.8} & \textbf{90.2} & \textbf{73.3} & \textbf{86.7}  \\
        %\midrule
        %\textbf{Ablation} \\
        \quad Prediction-based $\mathbf{e}_l$
         $\Rightarrow$ TF-IDF-based $\mathbf{e}_l$ & 91.8 & 94.4 & 81.6 & 90.0 & 71.5 & 85.9 \\
        %\midrule
        \quad $p_\theta(\mathbf{e}_l) > p_\theta(\mathbf{e}_w)$ Pairs 
        $\Rightarrow$ All Possible Pairs  & \textbf{92.9}  & 94.0 & 81.9 & 90.0 & 72.0 & 86.2\\
        %\midrule
        \quad $\mathbf{e}_l$ within Top-5
        $\Rightarrow$ Top-10 Predictions  & 92.5 & 94.0 & 82.1 & 89.6 & 72.6 & 86.2 \\
        \midrule
        \multicolumn{6}{l}{\textit{Models \textbf{without} Negative-aware Training}} \\
        \midrule
        GenBioEL~\cite{genbioel}  & 91.0 & 93.1  & 80.9 & 89.3 & 70.7 & 85.0\\
        \bottomrule
    \end{tabular}
    \caption{
    The ablation study on positive~($\mathbf{e}_w$) and negative~($\mathbf{e}_l$) pair selection during negative-aware fine-tuning.
    `$\Rightarrow$' indicates a modification in our method. 
    }
    \label{table:dpo_ablation}
\end{table*}

\section{Analysis}

\subsection{Ablation Study}

We conducted in-depth analyses on the selection of positive and negative pairs, the effect of similarity models on synonym retrieval, and the effect of pre-training. 
Additional analyses and results---including the number of synonyms used in positive-only training and the effect of optimization functions---can be found in Appendix~\ref{appendix:num_of_synonyms}.

\paragraph{Selection of positive and negative pairs}

\begin{table}[t!]
    \centering
    \footnotesize
    \begin{tabular}{lcc}
        \toprule
        \textbf{Model} & \textbf{NCBI} & \textbf{BC5CDR} \\
        \midrule
        TF-IDF (trigram-based) & \textbf{91.0} & \textbf{92.6} \\
        BioBERT-NLI & 90.1 & 71.9 \\ 
        SapBERT & 90.2 & 84.1 \\
        \bottomrule
    \end{tabular}
    \caption{Comparison of similarity models for retrieving positive and negative samples from KBs.}
    \label{table:similarity_models}
\end{table}

Analyzing the impact of how positive-negative pairs are constructed during negative-aware training is crucial for determining the optimal strategy for selecting hard negatives and the appropriate number of pairs. 
We investigated the effects of three factors: (1) negative sampling techniques (i.e., whether to use the model's incorrect predictions as negatives or rely on TF-IDF-based sampling), (2) the relative ranking of positive and negative samples, and (3) top-k selection (i.e., the number of negatives to include). 
Detailed results can be found in Table~\ref{table:dpo_ablation}. 
Ultimately, selecting negatives from the model's incorrect predictions proved to be the most important factor, with an average score difference of 0.8\%, while the other factors showed smaller differences of 0.5\%. 
More importantly, regardless of the specific negative-aware training approach, its application leads to significant performance improvements compared to models like GenBioEL, which do not incorporate such training. 
All models applying negative-aware training, including our ANGEL model, outperformed GenBioEL by 0.9\% to 1.7\% ($p \leq 1.7e^{-5}$ for all comparisons).

\paragraph{Effect of similarity models}

Similarity models play a critical role in retrieving synonymous terms from the KB, and their choice can have a substantial impact on overall system performance. 
To assess their effectiveness, we evaluated three models: (1) TF-IDF, (2) BioBERT-NLI,\footnote{\url{https://huggingface.co/gsarti/biobert-nli}} a sentence embedding model fine-tuned on natural language inference datasets, and (3) SapBERT. 
These models were incorporated into the positive-aware training framework of GenBioEL on the NCBI-Disease and BC5CDR datasets. 
As presented in Table~\ref{table:similarity_models}, the TF-IDF-based approach outperforms the two embedding-based models. 
Although the strong performance of trigram-based similarity highlights the utility of surface-level matching in BioEL, this does not imply that the task is inherently simple. 
While many synonyms exhibit similar surface forms, a substantial portion do not---posing challenging edge cases that demand more nuanced semantic understanding.

\paragraph{Effect of pre-training}

\begin{table}[t]
    \centering
    \footnotesize
    \begin{tabular}{lccc}
        \toprule
        \textbf{Model} & \textbf{FT} & \textbf{BC5CDR} & \textbf{AAP} \\
        \midrule
        BART & \xmark{} & 0.8 & 15.6 \\
        GenBioEL & \xmark{} & 33.1 & 50.6 \\
        \quad + \ours{} (\textbf{Ours}) & \xmark{} & \textbf{49.7} & \textbf{61.5} \\
        \midrule
        BART & \cmark{} & 93.0 & 88.7 \\
        GenBioEL & \cmark{}  & 93.1 & 89.3 \\
        \quad + \ours{} (\textbf{Ours}) & \cmark{}  & \textbf{94.5} & \textbf{90.2} \\
        \bottomrule
    \end{tabular}
    \caption{
    The top-1 accuracy of models with different pre-training strategies, along with the fine-tuned scores. 
    `FT' denotes fine-tuning, with~\xmark{}~representing pre-trained models without fine-tuning, and~\cmark{}~indicating models fine-tuned on human-annotated training sets.
    }
    \label{tab:pretraining}
    %\vspace{-3mm}
\end{table}

Table~\ref{tab:pretraining} highlights the effectiveness of \ours{}'s pre-training by comparing other pre-training methods. 
BART, pre-trained using a standard language modeling objective but not specifically tailored for BioEL tasks, shows the lowest performance.
In contrast, GenBioEL, pre-trained using synonyms from UMLS in a similar manner to our positive-only training, initially demonstrates a substantial performance advantage over BART. 
However, this gap narrows considerably after fine-tuning, to the point where it is no longer statistically significant.
When \ours{}'s negative-aware training is applied to GenBioEL, its performance improves significantly, achieving gains of 16.6\% on BC5CDR and 10.9\% on AAP. Even after fine-tuning, the performance gap remains noticeable, with a difference of 1.4\% on BC5CDR and 0.9\% on AAP.

\subsection{Understanding the Effectiveness of Negative-aware Training}

\begin{figure}[t!]
    \includegraphics[width=\linewidth]{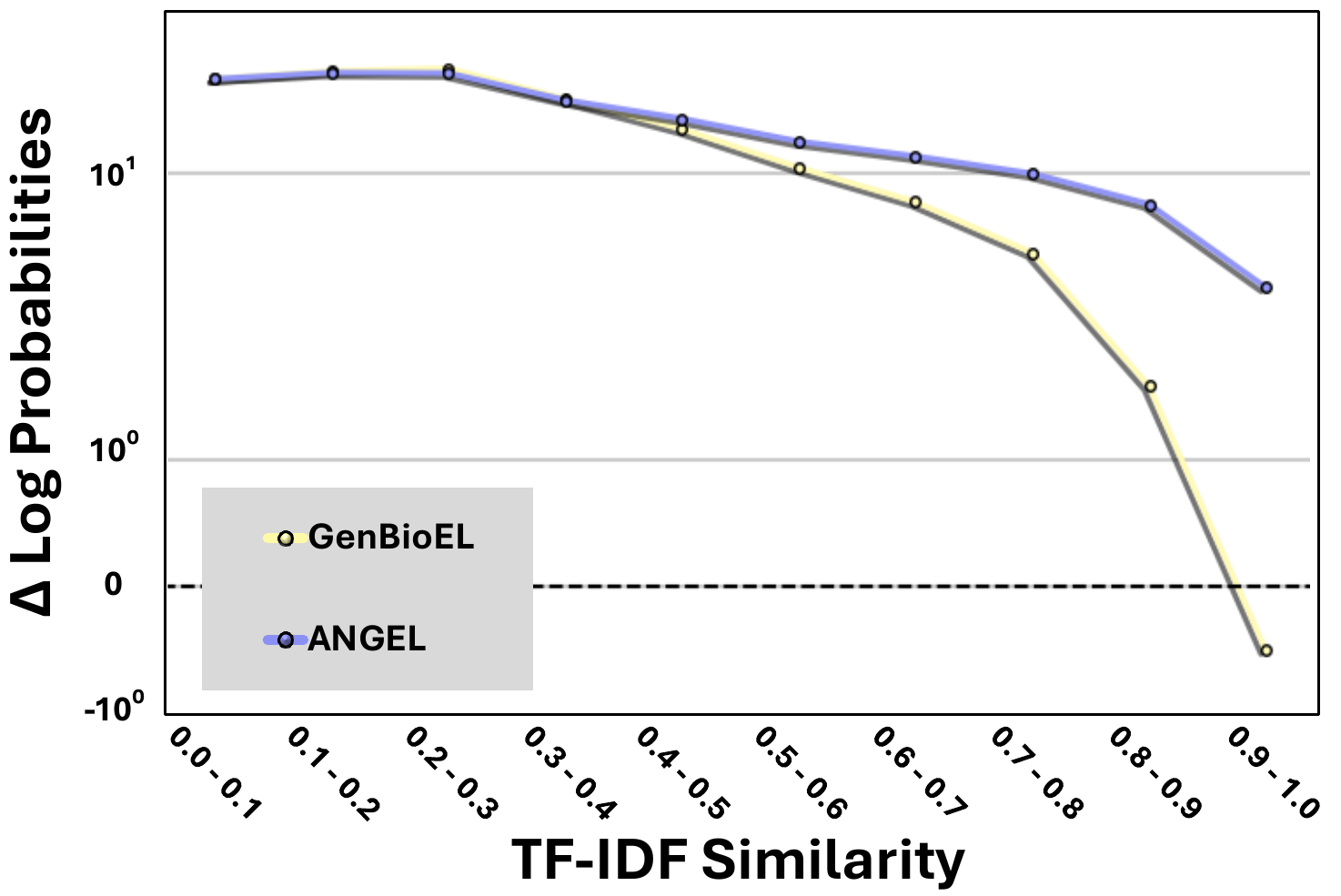}
    \caption{
    Analysis of the effect of negative-aware training. The x-axis represents the TF-IDF similarity between the input mentions and negative entities, while the y-axis depicts the difference in log probabilities between the top-1 positive prediction and negative entities for a given input mention.
    The NCBI-disease dataset was used.
    }
    \label{figure:logprob}
\end{figure}

Figure~\ref{figure:logprob} provide an interpretation of how negative-aware training leads to performance improvements.
While positive-only training increases the probability of identifying synonyms, it also raises the risk of incorrectly boosting the probability of negative samples that are morphologically similar to the input mention. 
In contrast, negative-aware training improves the identification of synonyms while simultaneously reducing the probability of incorrect negatives, making it particularly effective when these negatives share morphological similarity with the input mention.
To verify this, we divided input mention-negative pairs from the NCBI-disease dataset into 10 bins based on their tri-gram TF-IDF similarity.
We then computed the log probabilities of negatives within each bin for the corresponding input mentions, comparing them to the log probabilities of the top-1 positive predictions (i.e., the synonym assigned the highest probability by the model).
As the similarity between the input mention and the negative entities increased, the probabilities assigned by GenBioEL to positive and negative samples became more similar, eventually leading to higher probabilities for the negative samples. 
In contrast, our model demonstrated a clear distinction in behavior, consistently prioritizing positive samples over negatives.

\subsection{Error Analysis}
\label{subsec:error_analysis}

\begin{figure}[t!]
    \includegraphics[width=\linewidth]{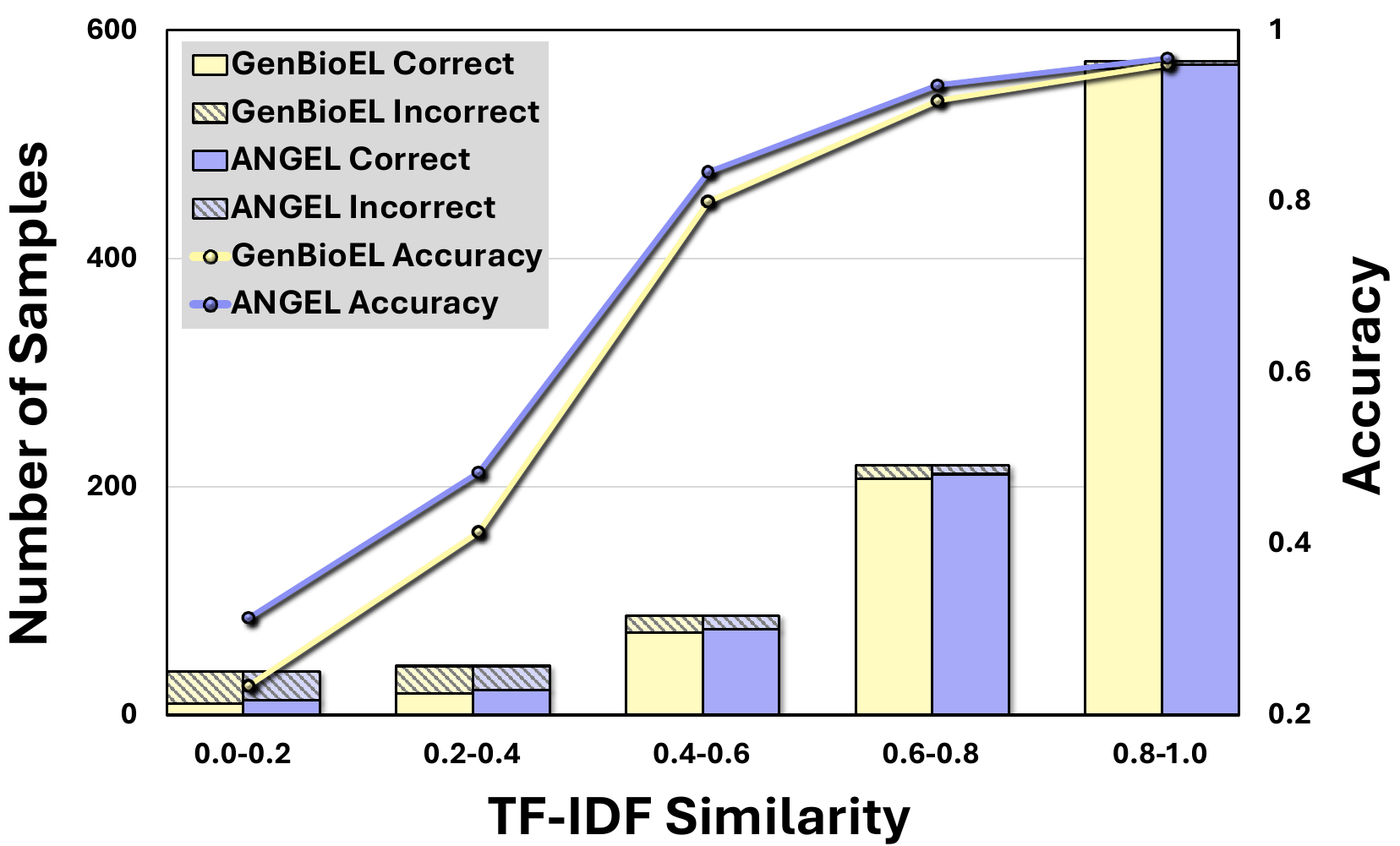}
    \caption{
    In-depth evalution of GenBioEL and our ANGEL models based on the TF-IDF similarity between the input mentions and gold-standard entities.
    The NCBI-disease dataset was used. 
    }
    \label{figure:tfidf}
    %\vspace{-3mm}
\end{figure}

\begin{figure*}[t]
    \includegraphics[width=\textwidth]{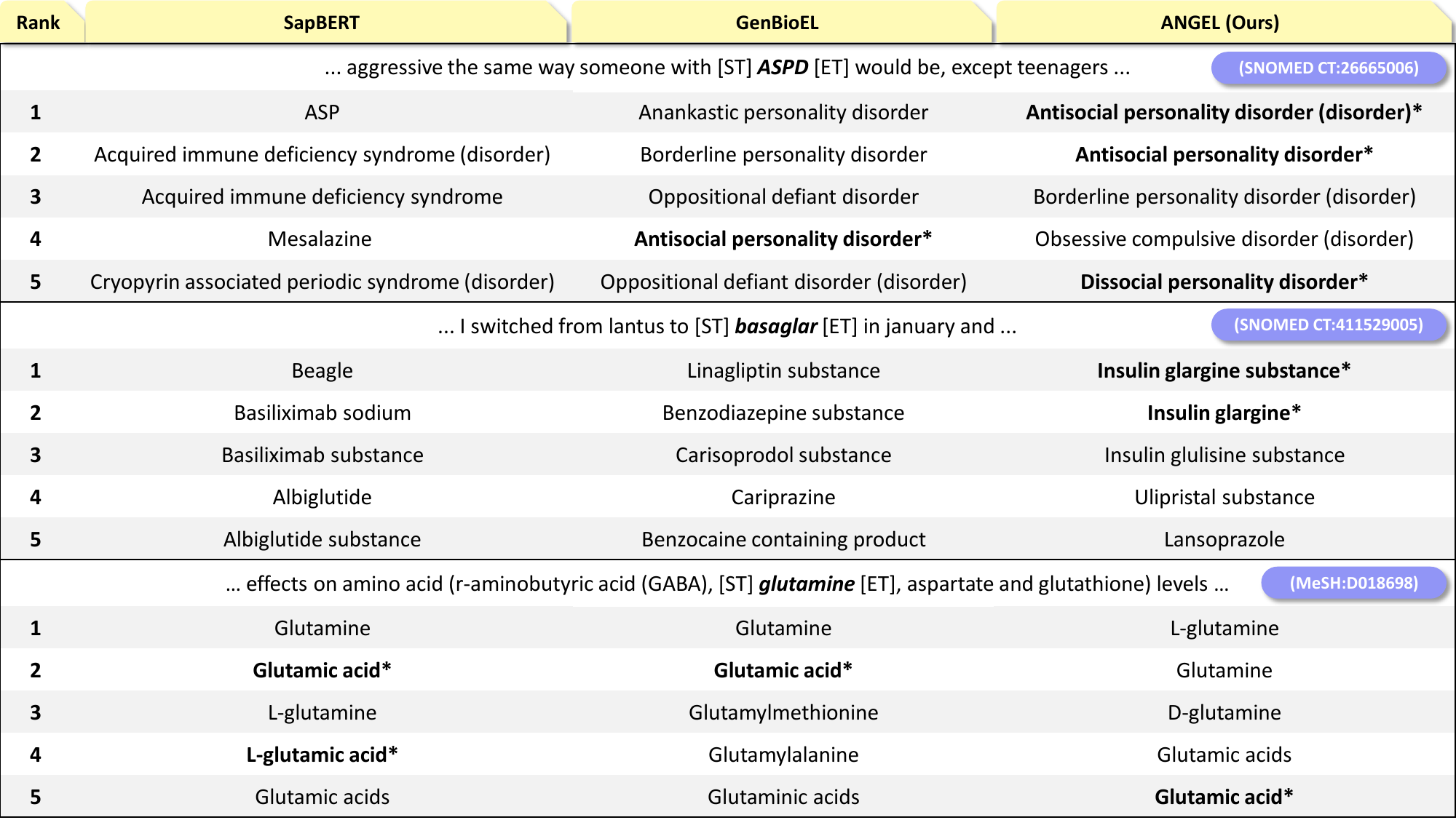}
    \caption{
    Top-5 predictions from different BioEL models are presented.
    Entity names with correct identifiers are highlighted in boldface with an asterisk.
    The first and second examples highlight the strengths of our model, while the final example illustrates its limitations. 
    For a detailed explanation, please refer to the main text.
    }
    \label{figure:case_study}
    %\vspace{-3mm}
\end{figure*}

We conducted an in-depth evaluation of the models based on the similarity between the input mentions and the gold-standard entities. 
Similarity was calculated using tri-gram TF-IDF, with the gold-standard entity determined as the candidate synonym with the highest similarity score to the input mention. The similarity scores, ranging from 0 to 1.0, were divided into five bins, and accuracy was measured for each bin.
As shown in Figure~\ref{figure:tfidf}, errors predominantly occurred in the 0-0.2 and 0.2–0.4 bins, as indicated by the height of the hatched bars, which represent the number of errors. 
This suggests that models tend to struggle when the surface forms of the input mentions are not closely aligned with those of the gold-standard entities.
Our method improves the generalizability of the model, leading to an overall reduction in GenBioEL's errors across all bins, with particularly notable improvements in cases of low similarity.
However, significant challenges remain, as the accuracy of our model is only 34.2\% in the 0–0.2 bin, highlighting the need for further improvement.

\subsection{Case Study}

Figure~\ref{figure:case_study} illustrates the predictions of SapBERT, GenBioEL, and~\ours{}. 
In the first example, the mention `ASPD' is an abbreviation for `antisocial personality disorder' (also known as `dissocial personality disorder'). 
SapBERT incorrectly predicts `ASP' due to the similarity in surface form. 
GenBioEL struggles to distinguish between correct entity names and those containing the words `personality disorder'.
In contrast, our model successfully identifies the correct entities, without being misled by false entity names that contain overlapping terms.
The second example involves the mention `basaglar,' a medication that contains insulin glargine, a long-acting insulin. 
The challenge here arises from the fact that product names can differ significantly from the biomedical terms used to describe their active ingredients. 
This discrepancy leads to failures in both SapBERT and GenBioEL, as they struggle to connect the brand name to its corresponding biomedical entity. 
Nevertheless, our model successfully identifies the correct entity, showcasing its ability to handle such complex cases effectively.
In the final example, our method was less effective. 
For the mention of `glutamine,' neither SapBERT nor GenBioEL identified the correct answer, but they did rank `Glutamic acid,' the correct entity, within the top 5 candidates.
Our model, however, ranked the correct answer slightly lower.
Consequently, while our model shows a notable improvement in top-1 accuracy, the increase in top-5 accuracy is relatively modest in some datasets. 
The effectiveness of our method also varies across different datasets. 
We discuss this limitation in more detail in Appendix~\ref{appendix:top-5}, noting that such cases are an area for further exploration.

\section{Conclusions}
In this study, we discussed the importance of negative samples in training generative BioEL models and introduced \ours{}, the first framework in this field to effectively incorporate negative-aware training into a generative model. 
Our models demonstrated the ability to learn subtle distinctions between entities with similar surface forms and contexts. 
Experimental results showed that \ours{} outperformed existing similarity-based and generative models, with notable performance improvements of 0.9\%, 1.4\%, and 1.7\% for BART, BioBART, and GenBioEL, respectively, achieving the best performance across five public BioEL datasets.

\section*{Limitations}

Our method is versatile and applicable to any generative model, but it has only been tested on encoder-decoder models and not on decoder-only models such as BioGPT~\cite{luo2022biogpt}.
We plan to further investigate the effect of our method on these models.
Additionally, it has not been tested on recent open-source large language models (LLMs)~\cite{touvron2023llama,chen2023meditron}. 
While we acknowledge that incorporating comparisons with LLMs and further assessing the effectiveness of our approach would be an interesting direction, using LLMs for entity linking presents new challenges.
The primary concern with larger models is their inefficiency, particularly regarding slower inference speeds and higher memory requirements, which may render them unsuitable for most real-world applications. 
This issue becomes particularly problematic in biomedical information extraction, where processing millions of publications to extract meaningful insights is essential.

Our negative-aware training method may not be limited to a specific domain, yet we have only evaluated it on biomedical-domain datasets, which restricts the demonstration of its broad applicability. 
Nevertheless, we would like to emphasize the reasons for focusing on the biomedical domain. Biomedical entity linking has unique characteristics that differentiate it from other domains, making this problem both challenging and interesting.
In general domains, ambiguity typically arises between different types of entities (e.g., whether ``Liverpool'' refers to a city or a sports club). 
Similarly, in the biomedical domain, ambiguity exists between different types, such as whether ``Ebola'' in Figure~\ref{figure:sub_figure} refers to a disease or a virus.
Additionally, biomedical entities often exhibit significant variations in their surface forms, even when they share the same identifier, i.e., they refer to the same entity. 
As shown in Figure~\ref{figure:case_study}, ``Basaglar'' can be expressed as other variations such as ``insulin glargine substance'' or ``insulin glargine.'' 
Furthermore, terms like ``\textit{substance}'' in the entity ``insulin glargine substance'' overlap with many other entities (e.g., ``Basiliximab \textit{substance},'' ``Linagliptin \textit{substance},'' ``Benzodiazepine \textit{substance}''), making the task even more complex.
Therefore, distinguishing between numerous candidates with similar surface forms is especially crucial in biomedical entity linking. We believe that our method, which trains the model using negative samples with similar structures, is particularly well-suited to tackle this challenge. However, exploring the application of our approach in other domains would be a valuable direction for future research.

\section*{Ethical Considerations}

This study complies with ethical standards, ensuring that all datasets and models adhere to their respective licenses and usage terms. 
While our method was evaluated on five widely used datasets, these serve primarily as benchmarks and may not fully capture real-world complexities. 
Although the model demonstrates significant improvements, its limitations in handling low-similarity cases highlight the need for thorough validation before deployment, particularly in sensitive applications.

\section*{Acknowledgements}
This research was supported by (1) the National Research Foundation of Korea (NRF-2023R1A2C3004176, RS-2023-00262002), (2) the Ministry of Health \& Welfare, Republic of Korea (HR20C002103), (3) ICT Creative Consilience Program through the Institute of Information \& Communications Technology Planning \& Evaluation (IITP) grant funded by the Korea government (MSIT) (IITP-2025-RS-2020-II201819).
M.S. was supported by (1) No. RS-2022-00155911: Artificial Intelligence Convergence Innovation Human Resources Development(Kyung Hee University), (2) No. RS-2024-00509257: Global AI Frontier Lab), and (3) the MSIT(Ministry of Science and ICT), Korea, under the ITRC(Information Technology Research Center) support program(IITP-2024-RS-2024- 00438239, 15\%).

%\clearpage
\bibliography{ANGEL}

\appendix
\clearpage
\renewcommand{\thetable}{\Alph{table}}
\renewcommand{\thefigure}{\Alph{figure}}
\setcounter{table}{0}
\setcounter{figure}{0}

\section{Datasets}
\label{appendix:datasets}

\begin{table*}[t]
    \centering
    \footnotesize
    %\resizebox{\textwidth}{!}{
    \begin{tabular}{lccccc}
        \toprule
        \textbf{Dataset} & \textbf{NCBI} & \textbf{BC5CDR} & \textbf{COMETA} & \textbf{AAP} & \textbf{MM-ST21pv} \\    
        \midrule
        \textbf{Entity types} & Disease & Disease/chemical & Medical concepts & Medical concepts & 21 UMLS types \\
        \midrule
        \multicolumn{6}{l}{\textit{\# Examples}} \\
        \textbf{Training} & 5,784 & 9,285 & 13,489 & 15,665 & 121,498 \\
        \textbf{Validation} & 787 & 9,515 & 2,176 & 793 & 40,600 \\
        \textbf{Test} & 960 & 9,654 & 4,350 & 866 & 39,922 \\
        \midrule
        \multicolumn{6}{l}{\textit{KB statistics}} \\
        \textbf{Entity names} & 108,092 & 809,929 & 904,798 & 3,398 & 6,051,091 \\
        \textbf{Identifiers} & 14,944 & 268,162 & 350,830 & 1,036 & 3,092,324 \\
        \bottomrule
    \end{tabular}
    %}
    \caption{The statistics of the benchmark datasets and their corresponding KBs.
    }
    \label{tab:dataset_stats}
\end{table*}

Table~\ref{tab:dataset_stats} presents the statistics of the five datasets used, along with their corresponding target knowledge bases. 

\paragraph{NCBI-disease} The NCBI-disease dataset~\cite{ncbi} contains 793 PubMed abstracts annotated with 6,892 disease mentions that are mapped to 790 unique disease concepts using the MEDIC ontology~\cite{medic}. 
MEDIC is a medical dictionary that integrates disease concepts, synonyms, and definitions from both MeSH~\cite{mesh} and OMIM~\cite{omim}, encompassing a total of 9,700 unique disease entities.
This dataset is primarily used for disease recognition and concept normalization tasks.

\paragraph{BC5CDR} The BC5CDR dataset~\cite{bc5cdr} includes 1,500 PubMed abstracts with 4,409 chemical entities, 5,818 disease entities, and 3,116 chemical-disease interactions. 
All annotated entities are mapped to the MeSH ontology~\cite{mesh}, which is a subset of UMLS~\cite{umls}. 
This dataset is widely used for biomedical entity recognition and interaction studies. 
To fit the purpose of our study, we use only the chemical and disease annotations and discard the interaction annotations.

\paragraph{COMETA} COMETA~\cite{cometa} focuses on layman medical terminology, compiled from four years of content across 68 health-related subreddits. 
This dataset consists of 20K biomedical entity mentions annotated with concepts from SNOMED CT~\cite{snomedct}. 
It is utilized for the normalization of consumer health expressions into standardized terminologies.

\paragraph{AskAPatient (AAP)} The AskAPatient dataset~\cite{askapatient} contains 8,662 phrases from social media language, each mapped to medical concepts from SNOMED CT~\cite{snomedct}. 
This dataset does not include contextual information, meaning that mentions are disambiguated solely based on the phrases themselves. 
Since the AskAPatient dataset lacks a test set, we employed a 10-fold cross-validation approach as outlined in the original paper by~\citet{limsopatham2016normalising}.
The statistics reported are the averages across these folds.

\paragraph{MM-ST21pv} The Medmentions dataset~\cite{medmentions} is a large-scale resource for biomedical entity recognition. 
The ST21pv subset includes 4,392 PubMed abstracts with over 200,000 entity mentions linked to 21 selected UMLS semantic types. 
This dataset provides a comprehensive resource for training and evaluating biomedical entity recognition systems. 
Unlike the original dataset, we use the 2020AA version of UMLS as the KBs because the 2017AA version of UMLS is not directly accessible. 
This leads to some differences after preprocessing due to variations between versions. 
Specifically, our dataset deviates from the original Medmentions dataset by 741 training samples (0.6\%), 284 validation samples (0.7\%), and 235 test samples (0.6\%).

\section{Hyperparameter Configurations}
\label{appendix:hyperparameters}

\begin{table*}[t]
    \centering
    \footnotesize
    %\resizebox{\textwidth}{!}{
        \begin{tabular}{l c c c c c c}
        \toprule
        \multirow{3}{*}{\textbf{Hyperparameter}} & \multirow{3}{*}{\textbf{Pre-training}} & \multicolumn{5}{c}{\textbf{Fine-tuning}} \\
        \cmidrule(lr){3-7}
        &  & \textbf{NCBI} & \textbf{BC5CDR} & \textbf{COMETA} & \textbf{AAP} & \textbf{MM-ST21pv} \\   
        \midrule
        \multicolumn{7}{c}{\textit{Positive-only Training}} \\        
        \midrule
        Training Steps & 80K & 20K & 30K & 40K & 30K & 40K \\
        
        Learning Rate & 4e-5 & 3e-7 & 5e-6 & 2e-5 & 5e-6 & 3e-5  \\
        
        Weight Decay & 0.01 & 0.01 & 0.01 & 0.01 & 0.01 & 0.01 \\
        
        Batch Size & 384 & 16 & 16 & 16 & 16 & 16 \\
        
        Adam $\epsilon$ & 1e-8 & 1e-8 & 1e-8 & 1e-8 & 1e-8 & 1e-8 \\
        
        Adam $\beta$ & (0.9,0.999) & (0.9,0.999) & (0.9,0.999) & (0.9,0.999) & (0.9,0.999) & (0.9,0.999) \\
        
        Warmup Steps & 1,600 & 0 & 500 & 1000 & 0 & 1,000 \\
        
        Attention Dropout & 0.1 & 0.1 & 0.1 & 0.1 & 0.1 & 0.1 \\
        
        Clipping Grad & 0.1 & 0.1 & 0.1 & 0.1 & 0.1 & 0.1  \\
        
        Label Smoothing & 0.1 & 0.1 & 0.1 & 0.1 & 0.1 & 0.1 \\
        
        \midrule
        \multicolumn{7}{c}{\textit{Negative-aware Training}} \\
        \midrule
        
        Epochs & 5 & 1 & 1 & 1 & 1 & 1 \\
        
        Learning Rate & 1e-5 & 1e-5 & 1e-6 & 5e-6 & 5e-6 & 5e-6 \\
        
        $\beta$ (DPO) & 0.1 & 0.1 & 0.1 & 0.1 & 0.1 & 0.1 \\
        
        Weight Decay & 0.01 & 0.01 & 0.01 & 0.01 & 0.01 & 0.01 \\
        
        Batch Size & 64 & 16 & 16 & 32 & 8 & 16 \\
        
        Warmup Steps & 1000 & - & - & - & - & - \\

        \bottomrule
        \end{tabular}
        %}
    \caption{Hyperparameters for positive-only training and negative-aware training.
    }
    \label{tab:hyperparameters}
\end{table*}

Table~\ref{tab:hyperparameters} details the hyperparameters used for positive-only training and negative-aware training across the BioEL benchmark datasets.
We searched for the optimal hyperparameter settings using the validation sets.
We refer to the study of~\citet{genbioel} to determine the range of the hyperparameters.
During pre-training, we used the same hyperparameters as in GenBioEL.
For positive-only training, we explored a range of training steps between 20K and 40K, a learning rate between 2e-5 and 3e-7, and batch sizes from 8 to 16, except during pre-training.
During negative-aware training, we fixed the $\beta$ at 0.1, in accordance with the basic configuration of DPO, and searched the hyperparameter space using a learning rate between 2e-5 and 1e-6 and batch sizes ranging from 8 to 64.
We used the source codes provided by \citet{genbioel}\footnote{\url{https://github.com/Yuanhy1997/GenBioEL}} and alignment handbook~\cite{alignment_handbook}\footnote{\url{https://github.com/huggingface/alignment-handbook}}.

\section{Ablation Study}
\label{appendix:num_of_synonyms}

\begin{table}[t]
    \centering
    \footnotesize
    \resizebox{\columnwidth}{!}{%
    \begin{tabular}{lcccc}
        \toprule
        \textbf{Model} & \textbf{FT} & \textbf{NCBI} & \textbf{COMETA} & \textbf{MM} \\
        \midrule
        BART & \xmark{} & 10.7 & 8.4 & 0.9 \\
        GenBioEL & \xmark{} & 58.2 & 42.4 & 10.4 \\
        \quad + \ours{} (\textbf{Ours}) & \xmark{} & \textbf{64.6} & \textbf{49.8} & \textbf{18.2} \\
        \midrule
        BART & \cmark{} & 90.3 & 80.4 & 70.1 \\
        GenBioEL & \cmark{} & 91.0 & 80.9 & 70.7 \\
        \quad + \ours{} (\textbf{Ours}) & \cmark{} & \textbf{92.8} & \textbf{82.8} & \textbf{73.3} \\
        \bottomrule
    \end{tabular}%
    }
    \caption{
    The top-1 accuracy of models with different pre-training strategies, along with the fine-tuned scores. 
    `FT' denotes fine-tuning, with~\xmark{}~representing pre-trained models without fine-tuning, and~\cmark{}~indicating models fine-tuned on human-annotated training sets.
    `MM' represents the MM-ST21pv dataset.
    }
    \label{tab:pretraining_2}
\end{table}

\paragraph{Effect of pre-training}
In addition to Table~\ref{tab:pretraining} which shows the effect of pre-training on BC5CDR and AAP, Table~\ref{tab:pretraining_2} demonstrates that ANGEL’s pre-training improves top-1 accuracy on NCBI-disease, COMETA, and MM-ST21pv, both before fine-tuning (\xmark{}) and after fine-tuning (\cmark{}).

\paragraph{Effect of optimization functions}

\begin{table}[t]
\centering
\footnotesize
\begin{tabular}{l c}
\toprule
\textbf{Model} & \textbf{Acc@1} \\
\midrule
GenBioEL                  & 85.0 \\
\midrule
ANGEL\textsubscript{FT} (Pairwise)   & 85.9 \\
ANGEL\textsubscript{FT} (CPO)~\cite{xu2024contrastive}        & 85.9 \\
ANGEL\textsubscript{FT} (SimPO)~\cite{meng2024simpo}      & 86.1 \\
ANGEL\textsubscript{FT} (DPO)~\cite{DPO}        & 86.2 \\
\bottomrule
\end{tabular}
\caption{Performance with different optimization functions. Average top-1 accuracy across the five benchmarks is reported.}
\label{tab:preference‐optimizers}
\end{table}

Our negative-aware framework is compatible with various optimization methods. To demonstrate this flexibility, we evaluated three additional loss functions during fine-tuning: (i) a simple pairwise loss, where $r_\theta(\mathbf{e} \mid \mathbf{x})$ is defined as $\log p_\theta(\mathbf{e}$;
and two preference optimization methods that build upon and improve DPO: (ii) Contrastive Preference Optimization (CPO)~\cite{xu2024contrastive}, and (iii) Similarity Preference Optimization (SimPO)~\cite{meng2024simpo}. 
Note that Equation~\ref{equation:general_loss} in the Method section presents a simplified version for clarity of explanation. 
In practice, CPO and SimPO introduce additional terms and require slight extensions to this formulation. 
For a detailed comparison of the exact equations, we refer readers to Table 7 in~\citet{meng2024simpo}, which provides a clear summary of the differences.

As shown in Table~\ref{tab:preference‐optimizers}, while DPO achieves the highest accuracy, the performance differences among the optimizers are relatively small. Regardless of the specific method used, all optimizers consistently enhance performance and outperform the GenBioEL baseline.

\paragraph{The number of synonyms}
To evaluate the impact of incorporating multiple synonyms during fine-tuning (Equation~\ref{equation:tfidf}), we conducted experiments by varying the number of synonyms associated with each mention, testing with 1, 3, and 5 synonyms. 
As a result, using 3 synonyms proved to be optimal, outperforming the approach that used only a single top-1 synonym in the study of \citet{genbioel}.

\begin{figure}[t!]
    \includegraphics[width=\linewidth]{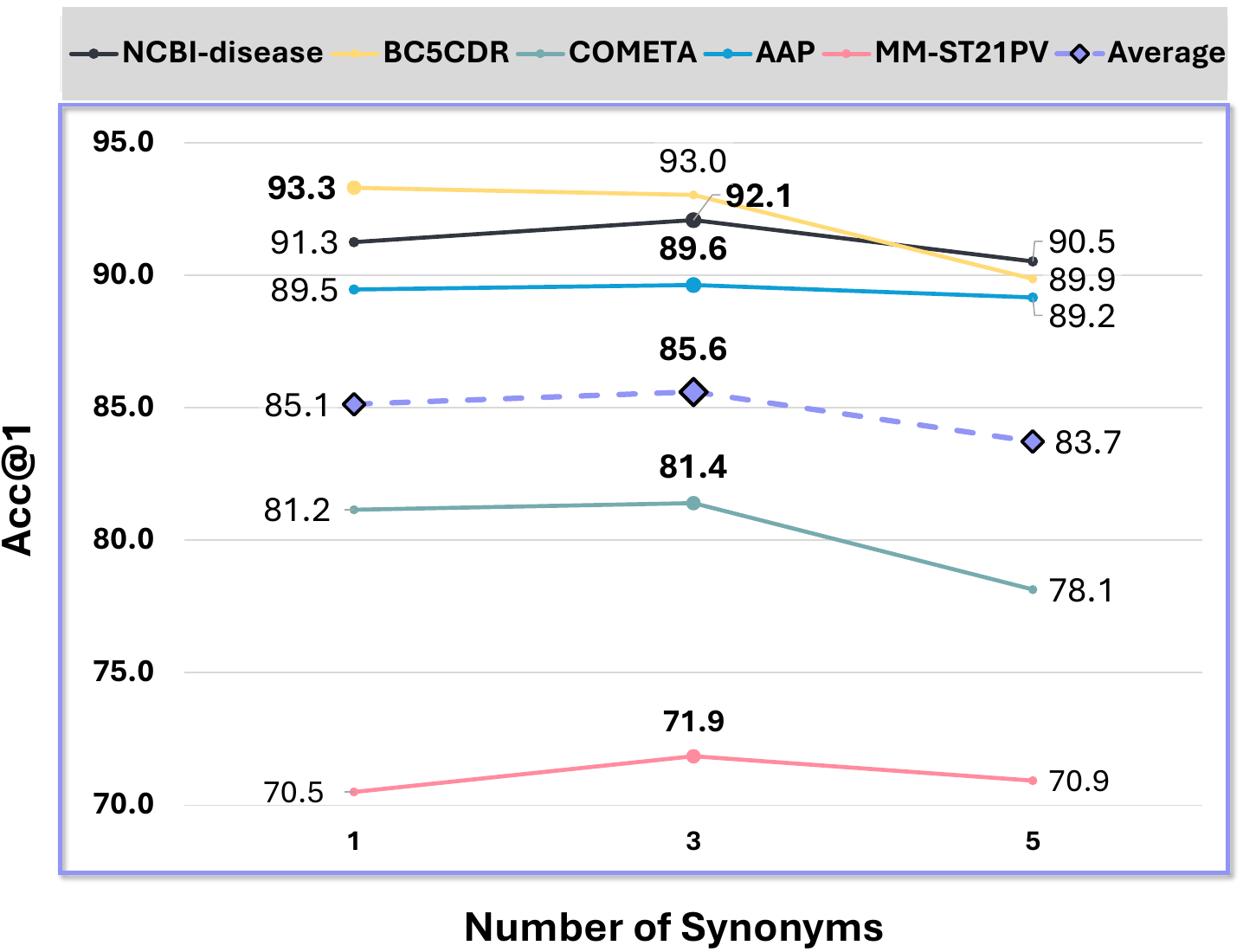}
    \caption{
    The ablation study to determine the optimal number of synonyms.
    GenBioEL with \ours{}$_\text{PT}$ was fine-tuned in this experiment.
    The scores are generally the highest when $k=3$.
    }
    \label{figure:syn_topk}
\end{figure}

\section{Error Analysis}
\label{appendix:error_cases_cometa}
\begin{figure}[t]
    \includegraphics[width=\linewidth]{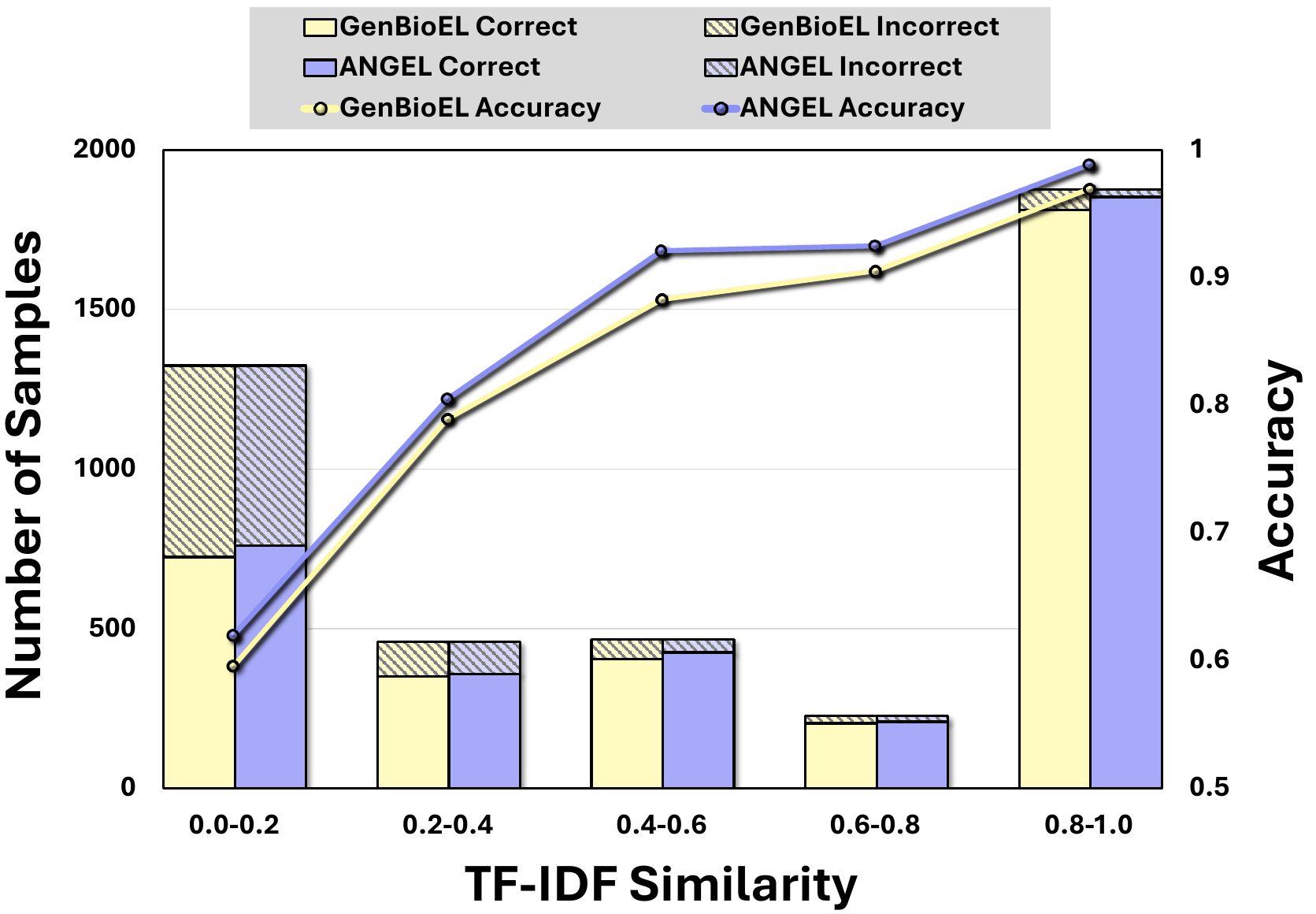}
    \caption{
    In-depth evaluation of GenBioEL and ANGEL using TF-IDF similarity between input mentions and gold entities on the COMETA dataset.
    }
    \label{figure:tfidf_cometa}
\end{figure}

\begin{figure}[t]
    \includegraphics[width=\linewidth]{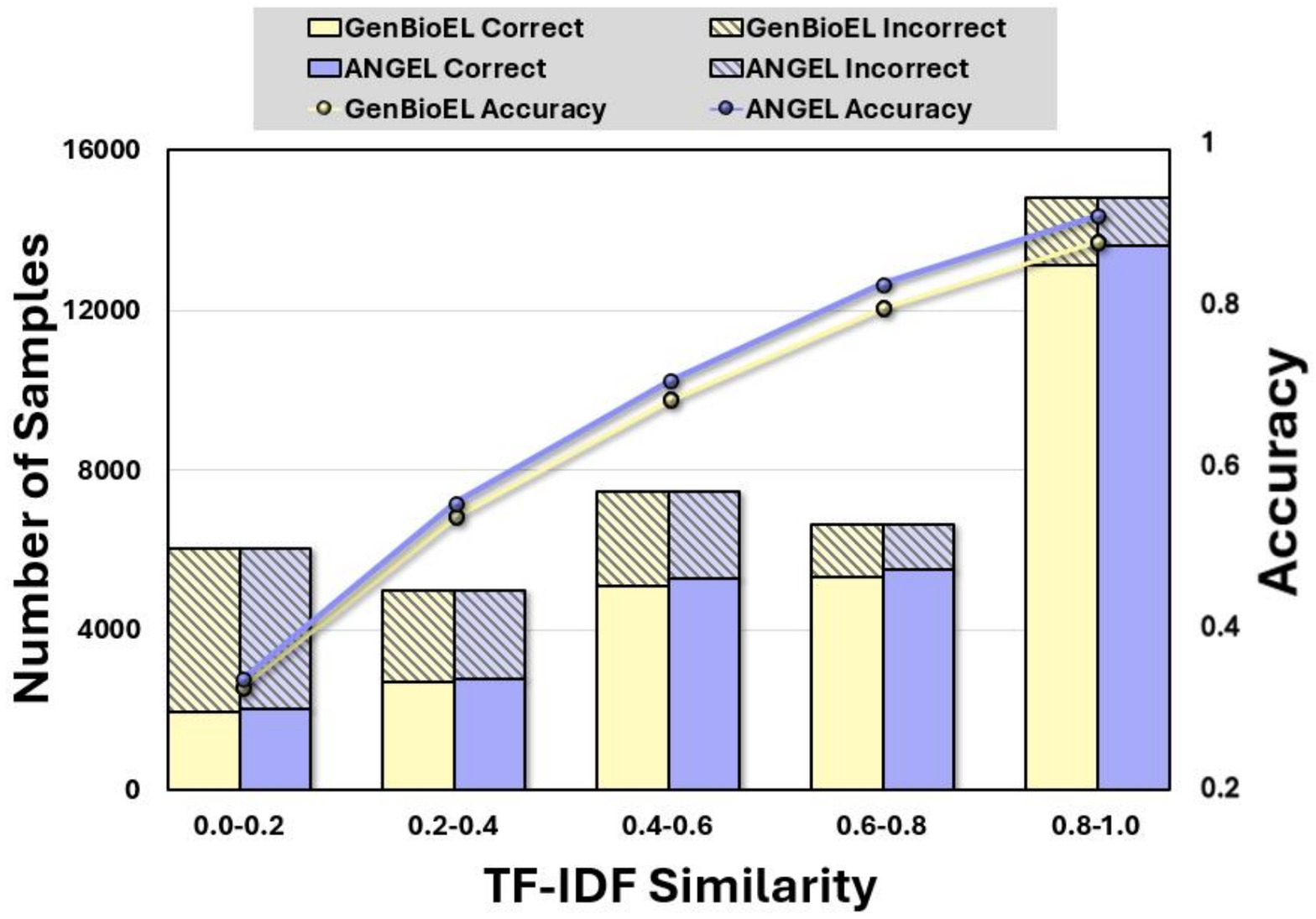}
    \caption{
    In-depth evaluation of GenBioEL and ANGEL using TF-IDF similarity between input mentions and gold entities on the Medmentions dataset.
    }
    \label{figure:tfidf_mm}
\end{figure}

Consistent with the analysis on the NCBI-disease dataset (Figure~\ref{figure:tfidf}), Figure~\ref{figure:tfidf_cometa} reveals that models on the COMETA dataset most frequently made errors in the 0.0–0.2 bin, where input mentions have low similarity to gold-standard entities. Across all similarity bins, our ANGEL framework consistently improved upon GenBioEL’s performance, leading to overall gains. A similar trend is observed in Figure~\ref{figure:tfidf_mm} for the MedMentions dataset, where ANGEL again outperforms GenBioEL across all bins. These results highlight the need for future research focused on reducing errors in low-similarity scenarios.

\label{appendix:error_cases_mm}

\section{Top-5 Accuracy}
\label{appendix:top-5}
\begin{table}[t]
    \centering
    \resizebox{\columnwidth}{!}{
    \begin{tabular}[width=\textwidth]{lcccc}
        \toprule
        \multirow{2}{*}{\textbf{Model}} & \multicolumn{2}{c}{\textbf{BC5CDR}} & \multicolumn{2}{c}{\textbf{AAP}} \\
        \cmidrule(lr){2-3} \cmidrule(lr){4-5}
        & \textbf{Acc@1} & \textbf{Acc@5} & \textbf{Acc@1} & \textbf{Acc@5} \\
        \midrule
        GenBioEL & 93.1 & 95.7 & 89.3 & \textbf{95.4} \\
        \quad + \ours{}$_\text{FT}$ & 94.4 & 96.5 & 89.5 & 94.7 \\
        \quad + \ours{}$_\text{PT + FT}$ & \textbf{94.5} & \textbf{96.8} & \textbf{90.2} & 95.2 \\
        \bottomrule
    \end{tabular}
    }        
    \caption{Comparison of top-1 and top-5 accuracy between the baseline model and models trained with~\ours{} method after fine-tuning and pre-training on the BC5CDR and AAP datasets.}
    \label{tab:acc5}
\end{table}

Table~\ref{tab:acc5} presents our model's top-1 and top-5 accuracy on the BC5CDR and AAP datasets. 
It compares the performance of our model in its baseline form (GenBioEL) and after fine-tuning (\ours{}$_\text{FT}$) and combined pre-training and fine-tuning (\ours{}$_\text{PT + FT}$).
Our approach consistently boosts top-1 accuracy across all datasets, though the trends in top-5 accuracy are less uniform. 
In BC5CDR, both top-1 and top-5 accuracy show significant improvements: top-1 accuracy rises by 1.4 percentage points (from 93.1\% to 94.5\%), and top-5 accuracy increases by 1.1 percentage points (from 95.7\% to 96.8\%). 
However, the AAP dataset exhibits a different pattern. 
While top-1 accuracy improves by 0.9 percentage points (from 89.3\% to 90.2\%), top-5 accuracy slightly declines: there is a 0.7 percentage points drop (from 95.4\% to 94.7\%) after fine-tuning and a 0.2 percentage points decrease (from 95.4\% to 95.2\%) after combined pre-training and fine-tuning.
This decline in top-5 accuracy may be due to the AAP dataset's limited contextual information, forcing the model to rely predominantly on the mention form, making it more challenging to maintain high accuracy across multiple predictions.
Additionally, the negative sampling strategy could unintentionally bias the model toward optimizing top-1 accuracy, thereby impacting top-5 performance.
In conclusion, while our method consistently improves top-1 accuracy, the occasional slight decreases in top-5 accuracy, as observed in the AAP dataset, underscore the need for further refinement to maintain balanced accuracy across different ranking levels.
Future work should focus on training strategies that preserve or enhance top-5 accuracy alongside top-1 improvements.

\end{document}